\documentclass[10pt,twocolumn]{article}

\usepackage[utf8]{inputenc}
\usepackage[T1]{fontenc}
\usepackage[margin=0.75in]{geometry}
\usepackage{times}
\usepackage{hyperref}
\usepackage{amsmath,amssymb}
\usepackage{fancyhdr}

\title{Aethon: A Reference-Based Replication Primitive for\\Constant-Time Instantiation of Stateful AI Agents}

\author{
Swanand Rao, Kiran Kashalkar, Parvathi Somashekar, Priya Krishnan\\
Next Moca Global, Inc.\\
\texttt{\{swanand, kiran, paru, priya\}@nextmoca.com}
}

\date{}

\fancypagestyle{plain}{
  \fancyhf{}
  \fancyfoot[C]{\footnotesize © 2026 Next Moca Global, Inc. Licensed under CC BY-NC-ND 4.0}

}

\begin{document}
\maketitle
\thispagestyle{plain}

\begin{abstract}
The transition from stateless model inference to persistent agent execution is beginning to reshape the systems assumptions that have governed applied artificial intelligence for more than a decade. Modern AI agents~\cite{ref36} are expected to do more than answer isolated prompts. They are expected to preserve continuity across interactions, accumulate context, invoke tools, collaborate with other components, and remain operable inside production workflows.

As the capabilities of large language models~\cite{ref1,ref2,ref3} have improved, these expectations have become technically plausible. What remains underdeveloped is the runtime substrate required to create, manage, and scale such entities without incurring severe latency, memory, and operational penalties. This paper introduces Aethon, a reference-based replication primitive for near-constant-time instantiation with respect to inherited structure.

The central idea is that an instance should not be treated as a fully materialized object that must be rebuilt from scratch each time it is created. Instead, an instance can be represented as a compositional view over a stable definition, layered memory, and local contextual overlays. By shifting instantiation from duplication to reference, Aethon turns creation into a lightweight operation whose cost is largely decoupled from the size of inherited configuration and memory.

We describe the conceptual model behind Aethon, explain the architectural decomposition required to support it, and analyze its implications for complexity, memory efficiency, personalization, and multi-agent orchestration~\cite{ref12,ref20}. We argue that this model is not merely an optimization technique. It is a different systems interpretation of what an executable agent instance actually is.

In doing so, Aethon points toward a class of runtime infrastructure in which such instances become cheap to spawn, safe to specialize, and practical to operate at production scale.
\end{abstract}

\section{Introduction}

Artificial intelligence systems are rapidly moving from a world of isolated inference requests to a world of persistent, interacting, stateful execution entities. In the earlier phase of applied machine learning, the dominant unit of computation was the prediction call. An input was passed to a model, an output was returned, and the system largely forgot the interaction.

With the rise of large language models~\cite{ref1,ref2}, a new pattern has emerged. The software layer above the model is no longer responsible only for packaging prompts and parsing answers. It is increasingly expected to sustain context, maintain role continuity, select tools, route work, decompose tasks, supervise execution, and preserve history across time.

The natural abstraction for this higher-order behavior is the AI agent~\cite{ref36}. An agent differs from a conventional model call not only in degree but in kind. It is not simply a longer prompt with memory attached. It is a software entity that participates in control flow. It may persist across sessions or appear transiently within a larger workflow. It may own a slice of state, inherit a behavioral policy, invoke external systems, or collaborate with sibling components in a distributed pattern of work~\cite{ref9,ref13}.

Once software systems begin to rely on such entities, infrastructure questions that were previously secondary become foundational. How should instances be created? How should they inherit structure? How should state be shared and isolated? How should personalized or task-specific branches be spawned without imposing severe per-instance overhead?

Many current systems answer these questions using a familiar pattern: explicit materialization. When a new instance is needed, the system loads a configuration, binds tool interfaces, reconstructs memory, allocates runtime state, and assembles an executable object. This is tractable when such entities are few, long-lived, and manually managed. It becomes increasingly problematic when systems evolve toward one-instance-per-user, one-instance-per-task, or highly dynamic multi-agent orchestration.

In such regimes, the cumulative cost of materialization becomes visible to end users, expensive to operators, and constraining to designers. The architecture itself begins to discourage the very patterns the application is trying to enable. The pressure is especially acute in systems that aspire to real-time responsiveness.

A user who asks an AI-enabled product to plan, retrieve, validate, execute, and synthesize across multiple subproblems expects a cohesive experience, not the latency profile of several cold starts disguised as intelligence. Likewise, a developer building an orchestration graph should be free to decompose a workflow according to semantic clarity, not according to fear of the infrastructure cost of spawning more execution identities. The current instantiation model frequently forces the opposite tradeoff.

It rewards flattening and reuse even when specialization would produce cleaner and safer behavior. The core observation motivating this paper is that runtime instances are rarely as unique, at the structural level, as current systems assume. The majority of what defines them is often stable or shared. Base role instructions persist across many invocations. Tool capabilities are commonly inherited from a common policy surface. Organizational knowledge, durable context, and operational constraints often apply to large families of instances rather than to one instance alone. Only a comparatively small portion of effective runtime state is genuinely local to a particular user, session, or task.

If that is true, then repeatedly reconstructing the entire instance is an artifact of an architectural assumption rather than a necessity imposed by the problem. Aethon begins with a different assumption. It treats an instance not as a newly assembled standalone object, but as a reference-oriented projection over shared structures and local deltas. Under this view, creating a new instance does not require duplicating all of its inherited substance. It requires creating a new identity that can be resolved, when needed, into an effective execution view.

This reframing changes the complexity profile of instantiation and the economic profile of multi-agent design. It also changes the language in which such systems can be described. Instead of building instances one by one, platforms can spawn them as needed from stable, governable definitions.

The argument of this paper is therefore twofold. The first claim is technical: reference-based replication can make stateful instance creation constant-time with respect to inherited structure, while preserving isolation through layered memory and copy-on-write semantics~\cite{ref34}. The second claim is conceptual: such a replication primitive is a better fit for the emerging ontology of production-grade agents than the default materialization model inherited from prior generations of software infrastructure.

If software is indeed entering an era in which these entities become commonplace execution primitives, then the infrastructure substrate must evolve accordingly.

\section{Background and Motivation}

The modern notion of the agent has emerged at the intersection of several independent trends. Language models have become powerful enough to serve as general reasoning engines over semi-structured tasks. Tool calling and function invocation frameworks have made it possible to connect those models to external software systems~\cite{ref5,ref6,ref40}. Workflow orchestration platforms have introduced compositional patterns for sequencing model calls, branching execution, and integrating retrieval~\cite{ref12,ref22,ref27,ref28}. At the same time, product expectations have shifted. Users increasingly want systems that feel continuous, contextual, and proactive rather than merely responsive.

These trends together have made the agent abstraction practical rather than speculative. Yet the surrounding runtime assumptions still reflect an earlier era. In classic service-oriented software, the dominant optimization problem was handling large volumes of independent requests efficiently. Statelessness was an advantage, not a limitation, because it simplified scaling and fault isolation. Model serving inherited much of this logic. The system optimized token throughput, latency per request, hardware utilization, and batching behavior. Even applications that layered light memory on top of models often maintained a fundamentally request-response architecture.

The new generation of agentic systems stresses that architecture because the meaningful unit of behavior is no longer only the request but the persistent execution identity that survives or recurs across requests. The infrastructure challenge becomes clear when such applications are examined not as demos but as operational systems.

Consider a customer-support environment in which every active ticket may invoke a dedicated issue-resolution component. Consider a software platform in which every user receives a personalized workspace entity, and every major task decomposition spawns supporting branches for retrieval, validation, execution, and summarization. Consider a sales or RevOps environment in which each account receives a scoped execution layer that acts on its own context while inheriting organizational policy, product knowledge, and CRM permissions. In each of these settings, creation is not exceptional. It is part of the normal flow of the system. Any inefficiency in instantiation is multiplied by the scale and cadence of usage.

Materialization-based approaches struggle here for several reasons. First, they couple creation cost to inherited size. If a base definition accumulates rich instructions, broad tool access, and large context layers, every descendant becomes expensive to initialize even if it needs only a tiny local specialization. Second, they blur the line between sharing and isolation. Developers either duplicate too much, wasting resources, or reuse too aggressively, risking state leakage and unclear ownership boundaries. Third, they make lineage harder to preserve. Once an instance has been deeply copied and rehydrated into an opaque runtime object, reconstructing what it inherited and how it diverged becomes more difficult.

These limitations lead teams toward compensating behaviors. One common behavior is over-reuse: systems keep a smaller number of long-lived instances alive and route many tasks through them. This reduces setup cost but can create muddled identity, hidden context carry-over, and complicated concurrency semantics. Another behavior is over-flattening: designers compress multiple roles into one broad prompt because instantiating separate specialized branches feels too expensive. This can make prompts brittle and increase the cognitive burden on both the model and the operator. A third behavior is anticipatory preloading: systems keep pools of partially initialized instances in memory to reduce apparent latency, effectively trading runtime responsiveness for background waste and operational complexity.

The deeper problem is not that these workarounds are unsophisticated. It is that the base abstraction is misaligned with the emerging use case. If such entities are expected to branch, specialize, and proliferate, then the architecture should make those operations cheap and explicit. In many other areas of computing, systems became more powerful not when individual runtime objects became more manually optimized, but when the unit of creation became light enough to enable new patterns. Virtual memory, fork semantics, copy-on-write pages, immutable data structures, and lazily evaluated expressions all expanded the design space because they weakened the assumption that new identity requires full duplication~\cite{ref14,ref15,ref16,ref17,ref34}.

Aethon is motivated by the claim that runtime instantiation in this setting deserves the same kind of rethinking. Much of effective behavior can be expressed as the composition of a stable definition, inherited context, and a relatively small number of local deltas. If that is so, then a new instance should often be representable by a new reference rather than a new copy.

The opportunity is especially significant because such systems sit at the junction of cognition, memory, control flow, and policy. A more efficient instantiation primitive does not merely save cycles. It changes what kinds of systems are practical to build.

The motivation for Aethon is therefore strategic as much as technical. It aims to enable a mode of software design in which specialization is easy, multi-agent decomposition is natural, personalization is scalable, and lineage is preserved through explicit inheritance. In other words, it attempts to supply a missing primitive for the operating model of production AI agents.

\section{Aethon Conceptual Framework}

Aethon rests on a straightforward but far-reaching conceptual distinction: the difference between an instance as a materially assembled object and an instance as an effective runtime view. In the first interpretation, which dominates most existing systems, a new entity exists only once the system has fully constructed it. In the second interpretation, it can be said to exist once the system has created a stable identity through which eventual behavior can be resolved.

The latter interpretation is what Aethon makes operational.

The concept becomes clearer when identity is decomposed. Let $D$ denote a canonical definition that specifies the durable structure of a class, including role semantics, instructions, capabilities, and policy constraints. Let $M_s$ denote one or more layers of shared memory or inherited context. Let $M_i$ denote local state unique to a particular instance or branch. Let $C_i$ denote contextual bindings such as user scope, task metadata, or session-local parameters.

An effective instance $A_i$ can then be interpreted as a composition over these layers rather than as a self-contained monolith:

\[
A_i = f(D, M_s, M_i, C_i)
\]

where $f$ is a composition operator rather than a one-time constructor.

The important implication is that an instance does not need to physically own all of $D$, $M_s$, $M_i$, and $C_i$ in flattened form in order to function. What it needs is a stable reference to the relevant inherited structures plus the local information that determines how those structures should be resolved at execution time.

Aethon therefore represents an instance through a reference record that points into the stable substrate and captures local deltas and lineage metadata. The representation is not valuable because it is small in the abstract. It is valuable because it mirrors the actual distribution of sameness and difference among related execution identities.

This approach introduces the idea of deferred realization. Under Aethon, creating an instance does not necessarily mean paying the full cost of assembling every inherited component into an eagerly materialized structure. Instead, the system can defer composition until execution or until a later boundary at which the effective state is truly needed. This differs from simple caching. The system is not just postponing an expensive rebuild. It is acknowledging that the rebuild may not be the right semantic operation in the first place.

If most inherited structure remains shared and only a narrow slice must be individualized, then flattening everything eagerly may be both wasteful and misleading. Deferred realization offers another subtle benefit: freshness. In environments where context, policies, or shared memory layers may change over time, eagerly materialized descendants can become stale unless they are refreshed or rebuilt. Aethon allows a descendant reference to remain lightweight while resolving the relevant inherited state at execution time, subject to versioning and access rules.

This does not mean all inherited structures are mutable in place. In many cases, definitions should remain version-stable for reproducibility. Rather, it means the system gains more flexibility in deciding which layers are fixed lineage anchors and which are dynamically resolved contextual dependencies.

Aethon also turns lineage into a first-class property. If one instance descends from another through specialization, scoping, or branching, that relationship can be preserved explicitly in the reference graph rather than erased through deep copying. This matters because lineage is operationally useful. It supports provenance, auditability, debugging, and rollback.

It also allows orchestration logic to reason more directly about families of related instances. For example, a supervisory branch might spawn several descendants that each inherit the same structural definition but receive distinct task scopes and local overlays. In a materialization model, these descendants quickly become opaque siblings. In Aethon, they remain traceable as branches of a common lineage.

Another consequence of the framework is that specialization becomes cheaper and more expressive. A system can derive an execution identity for a particular subtask not by cloning the entire parent but by defining a narrower view over the parent's inherited structures. That narrower view may restrict tools, add a temporary role modifier, attach a task-local memory overlay, or bind a different output contract. The important point is that specialization is expressed through the reference and resolution mechanism rather than through duplicated construction. This makes the semantics of specialization clearer and the infrastructure cost lower.

The conceptual framework of Aethon therefore invites a broader reinterpretation of agents. Instead of treating them as heavyweight runtime objects that happen to carry state, it treats them as compositional execution identities whose behavior arises from layered inheritance and scoped deltas. Once such entities are seen this way, near-constant-time instantiation with respect to inherited structure becomes not just plausible but natural.

\section{System Architecture}

Aethon's architecture is designed to preserve the separation between stable definition, inherited state, local divergence, and execution-time realization. This separation allows the system to avoid flattening all structure into each new instance. At a high level, the architecture can be understood as comprising a definition substrate, a reference substrate, and a resolution substrate, all operating over a layered memory model with explicit lineage and scope semantics.

The definition substrate is responsible for representing durable structure. This includes role identity, behavioral instructions, declared capabilities and tool access policies, interface contracts, and structural metadata required for introspection. The key property of the definition substrate is stability. Definitions may evolve over the life of a platform, but a given versioned definition serves as a stable anchor for descendant references. This is crucial because structural sharing requires a trustworthy object of inheritance. If the base were mutated implicitly beneath descendants, the semantics of lineage would become unstable and reproducibility would suffer.

The reference substrate is where instantiation becomes lightweight. A reference is the minimal execution-capable identity for a derived instance. It records which canonical definition the instance descends from, which inherited memory layers are in scope, what local overlays or deltas apply, and what lineage or version metadata are necessary for interpretation. It may also carry scoping constraints, such as user boundaries, task restrictions, or limited tool surfaces.

The reference is not merely a pointer in the narrow sense. It is a semantic handle that captures how a descendant should be understood relative to the shared substrate.

The value of this design becomes evident when systems need to generate many related instances. A customer-specific descendant, a workflow-specific descendant, and a task-branch descendant may all share the same root definition while differing only in overlays and scope. In a materialization model, each of these would be reconstructed into a separate object. In Aethon, each becomes a new reference over the same stable base. Instantiation cost is therefore tied primarily to the cost of creating the reference record and registering lineage, not to the cost of duplicating the full inherited structure.

The resolution substrate turns references into effective execution state. When an instance is about to act, the resolver consults the reference, retrieves the applicable definition, composes the relevant inherited memory layers, applies overlays, enforces scope constraints, and produces the effective view from which execution proceeds. Importantly, the resolver need not flatten everything into a persistent standalone object. It can construct a transient execution view in which shared layers remain shared and local layers are introduced only where necessary.

This allows the system to preserve both efficiency and explicit lineage. Architecturally, this means that execution is no longer the same thing as instantiation. The two are decoupled. Instantiation becomes the cheap creation of a new semantic identity, while execution becomes the context-sensitive realization of that identity.

Another important feature of the architecture is versioned compatibility. Because the definition substrate is explicit and stable, systems can support controlled evolution of definitions without breaking descendant lineage. A new version can be introduced while existing descendants continue to point to the prior version from which they were created. This allows organizations to improve logic over time without sacrificing auditability. It also creates a natural foundation for rollback and comparative evaluation, since the architecture retains a clear map from descendant behavior to ancestral structure.

The architecture is equally relevant to enterprise control requirements. In production settings, it is often necessary to know not only what a system did, but what it was allowed to do and what it inherited at the time of execution. By making references, definitions, and memory layers explicit architectural entities, Aethon creates a more inspectable substrate than one in which state is flattened and opaque. Access control and operational policy can be enforced at the level of inheritance and scope rather than only at the level of broad application permissioning.

Aethon's system architecture therefore supports more than performance. It supports a disciplined model of identity, inheritance, and execution that scales operationally and remains legible under real production pressure.

\section{Memory Model}

Memory is one of the central reasons these systems are valuable, and it is also one of the central reasons they are difficult to scale. An execution identity that cannot preserve context across time is often little more than a stateless assistant with extended prompting. But one that preserves too much state too eagerly or too redundantly becomes operationally expensive, difficult to isolate, and hard to control.

Aethon's memory model is designed to avoid that tradeoff by separating memory according to scope and by allowing inheritance without compulsory duplication.

The first principle of the model is that memory is layered, not monolithic. Different kinds of context live at different semantic scopes. Some information belongs at an organizational or platform level. This may include domain policies, durable operational conventions, or long-lived shared facts that many descendants should be able to reference. Other information belongs at the level of a family or lineage. This could include specialized behavioral priors, task-type memory, or reusable context associated with a category of work. Still other information belongs to a specific user, account, session, or workflow. Finally, there is highly local state that arises only within a particular execution branch or temporary descendant.

Aethon treats these as distinct memory layers rather than forcing them into a single duplicated bundle. This layering matters because not all memory should be handled the same way. Shared organizational memory may be relatively stable and broadly visible. User memory may need stricter boundaries and privacy controls. Task-local state may be ephemeral and disposable. If all of these layers are flattened into each new instance, the system loses both efficiency and semantic clarity.

Aethon instead treats effective memory as the result of structured composition across layers. The instance sees the memory environment relevant to its scope, but the underlying storage and inheritance remain segmented.

The second principle is copy-on-write semantics~\cite{ref34}. Structural sharing is desirable for both space efficiency and instantiation speed, but sharing cannot come at the expense of isolation. If two descendants both read from the same inherited memory layer, there is no need to duplicate that layer merely because both exist. If one of them must diverge through mutation, however, the system must preserve that divergence locally without corrupting siblings or ancestors.

Copy-on-write provides this behavior. The inherited layer remains shared until a write requires a local overlay or detached copy. The result is a memory model in which duplication tracks actual divergence rather than potential divergence.

This has powerful consequences in environments with many related descendants. Suppose a workflow spawns multiple specialized branches that all require access to the same customer context and organizational policies, but only one of them generates a task-specific intermediate hypothesis that should remain local. Under a materialization model, all descendants may receive full copies of the inherited context plus their own local state. Under Aethon, the inherited context can remain shared and only the task-local mutation needs its own overlay. Space consumption therefore grows with meaningful divergence rather than with the raw number of descendants.

The memory model also supports lineage-aware introspection. Because local overlays arise relative to inherited layers, it becomes possible to ask not only what memory an instance used, but what it inherited and where it diverged. This is crucial for debugging and auditability. A surprising behavior may not be caused by local state alone; it may emerge from a subtle interaction between inherited organizational context, lineage-specific memory, and a task-specific overlay. A flat snapshot can obscure such relationships. A layered model preserves them.

Another important feature of the memory model is compositional resolution. When an instance is invoked, the system determines the effective memory environment by consulting the relevant layers in scope and applying precedence rules. The result is similar in spirit to variable resolution across lexical scopes or environment stacks in programming languages, though the semantics are broader because memory may include semi-structured facts, prior interactions, and policy surfaces rather than only symbolic bindings. The key point is that memory is assembled in an ordered way rather than copied wholesale.

Personalization benefits directly from this model. A system may want every user to have a dedicated contextual intelligence surface while still inheriting shared product knowledge, access rules, and enterprise memory. A naive implementation would duplicate much of the upstream substrate for each user. Aethon allows personalization to be expressed through comparatively small user-level overlays on top of shared layers. This makes it far more feasible to support many personalized descendants at once without turning memory duplication into the dominant systems cost.

From an operational control standpoint, the layered memory model is equally valuable. Different layers can be governed differently. An enterprise memory layer may be centrally managed and durable. A user-local layer may be subject to privacy restrictions or retention limits. A task-local layer may be short-lived and auditable. By keeping these layers explicit, Aethon enables more fine-grained access control and lifecycle policy than a flat memory bundle would allow.

The memory model is therefore not an accessory to Aethon's instantiation claim. It is a prerequisite. Constant-time instantiation becomes useful only if inherited state can remain inherited after creation. The layered, copy-on-write memory model is what makes that possible while preserving the isolation, traceability, and personalization properties that production systems require.

\section{Complexity and Performance Characteristics}

The central quantitative claim of Aethon is that instance creation can be made constant-time with respect to the size of inherited structure. This claim is best understood as a statement about what work belongs to the act of creation. In a materialization-based architecture, creation is tightly coupled to reconstruction. The system loads or copies configuration, reconstructs memory, rebinds capabilities, and produces a standalone runtime artifact. If the inherited structure grows, the creation cost tends to grow with it.

In Aethon, creation is redefined as the production of a semantic reference into shared substrates. The inherited structure remains where it is. The new work consists primarily in producing a new identity, registering lineage, and recording the local delta and scope.

To make this distinction concrete, consider a conventional runtime entity whose effective behavior depends on a definition of size $d$, an inherited memory environment of size $m$, and a set of bindings or capabilities of size $b$. A materialization-oriented system will often pay a cost that is some function of $d$, $m$, and $b$ each time a new instance is assembled. Even if caching reduces constants, the semantic contract of creation still involves reproducing substantial inherited substance.

In contrast, an Aethon-style system can create a reference whose size is largely independent of $d$, $m$, and $b$, provided the reference need only point to those inherited structures rather than duplicate them. Under that assumption, creation cost is effectively $O(1)$ in the size of inherited state.

This does not mean that all associated work disappears. Some costs are shifted to resolution time, and some systems may choose to pre-resolve certain layers opportunistically for performance reasons. But the shift matters enormously at scale. In systems where agents~\cite{ref36} are created frequently and many descendants share common ancestors, paying full inherited cost at each instantiation quickly becomes untenable. By contrast, paying near-constant cost for identity creation while deferring or minimizing realization allows systems to create more specialized branches within the same latency and resource envelope.

The same principle changes the memory growth curve. In a naive duplication model, total memory tends to grow roughly with the number of instances times the amount of inherited state each one carries. Deduplication mechanisms can reduce this in practice, but the conceptual model still begins with broad reconstruction. Aethon's layered inheritance and copy-on-write behavior instead make memory growth track divergence rather than population. Shared memory layers are stored once. Descendants incur additional space primarily when they introduce local overlays. As a result, workloads with high structural commonality and relatively small local mutations can exhibit much more favorable space scaling.

From an orchestration standpoint, this changes decision-making. In a materialization-heavy environment, every branch in a workflow has an implicit infrastructure price. A planner or workflow designer must ask not only whether a specialized branch is semantically appropriate, but whether the cost of creating it is worth paying. Such cost pressure subtly biases systems toward broader, more overloaded execution identities. Under Aethon, the marginal cost of spawning a descendant drops significantly. This makes specialization and decomposition more attractive as routine design patterns.

There is also a latency implication for user experience. When an interactive system must respond to a request by creating task-specific subcomponents, the user perceives not only model latency but orchestration latency. If subcomponent creation is expensive, the system feels hesitant even before the model begins substantial reasoning. If creation is cheap, the orchestration layer can afford to behave more dynamically without exposing obvious setup penalties.

Aethon's performance characteristics are therefore best understood as architectural leverage. The goal is not to claim that every downstream operation is cheap. Model inference remains expensive. External tools remain variable. Memory resolution must still be implemented carefully. The important change is that system proliferation no longer forces the platform to pay the full cost of inherited structure at each point of creation.

\section{Multi-Agent Orchestration}

The case for Aethon becomes strongest when one considers the likely trajectory of agentic software. The future does not appear to point only toward larger and more monolithic execution identities. It points toward richer orchestration among many narrower, context-specific descendants that can be spawned, supervised, and recombined within a workflow.

Such systems demand an instantiation primitive that behaves more like process spawning than artisanal reconstruction. Aethon aims to provide exactly that.

Multi-agent systems~\cite{ref20} derive their appeal from modularity. A planning component can delegate retrieval to one descendant, verification to another, and execution to a third. A user-facing assistant can create temporary subcomponents specialized to a single tool surface or task family. A supervisory layer can fan out work across multiple branches and reconcile their outputs before making a final decision.

These designs improve clarity because each descendant can be scoped more narrowly and given more precise instructions or permissions. They also improve fault isolation because behavior is distributed across semantically distinct units rather than merged into one overburdened role. Yet the practical adoption of such patterns is often limited by the cost of creating descendants. If every fan-out event incurs heavy initialization, developers are incentivized to collapse roles together.

Aethon reduces that mismatch. By making descendant creation lightweight, it allows semantic decomposition to become the main criterion for branching rather than infrastructure overhead.

Lineage is especially valuable in multi-agent contexts. A retrieval descendant may inherit customer scope and product policy from its parent while adding a narrow task overlay and restricted tool surface. A validation branch may inherit the same base but operate under a different decision threshold and output contract. Because these descendants are created through references rather than deep copies, the system can preserve explicit ancestry. This makes it easier to inspect which common structures shaped them and how each one diverged locally.

Aethon also improves the economics of experimentation in orchestration. Teams building such workflows often discover useful structure only by trying many decompositions. If creating new specialized descendants is expensive, experimentation becomes operationally costly and psychologically discouraged. If creating them is cheap, designers can explore richer graphs more freely.

Another advantage concerns constrained collaboration. In enterprise settings, not every descendant should inherit the full capability surface of its ancestor. Some should read but not write. Some should retrieve but not execute. Some should see user-level context without seeing broader organizational state. Aethon's reference and resolution model makes these restrictions easier to express because descendants can be defined as scoped views rather than as fully copied entities with permissions manually stripped after the fact.

There is a further operational benefit in systems that require high concurrency. A customer support platform may need to run many resolution branches in parallel. A sales intelligence system may need to produce scoped account-level analyses across many active opportunities simultaneously. A product workflow may require dozens of short-lived descendants inside a single user interaction. In all such cases, the ability to create many stateful descendants without linear setup growth is not merely convenient. It is the difference between an architecture that scales and one that constantly fights its own orchestration ambitions.

The broader implication is that multi-agent systems should not be treated as an exotic extension of single-agent design. They are likely to become a normal pattern in production software. If so, then the infrastructure substrate must support cheap, lineage-aware branching with explicit scope control. Aethon offers a candidate primitive for that substrate by making descendants inherit through reference, localize through overlays, and execute through runtime resolution.

\section{Governance, Isolation, and Enterprise Implications}

As these systems move from experimentation into enterprise production, control requirements become inseparable from architecture. Organizations do not merely want systems that are capable. They want systems that are inspectable, auditable, reproducible, and compatible with the boundaries through which real businesses manage risk.

This is one of the reasons Aethon's reference-based model matters beyond performance. By making inheritance, lineage, and scope explicit, it creates a more controllable substrate for production operation.

In many enterprise contexts, the most serious operational questions are not about raw model quality. They are about where context came from, what the system was allowed to access, which version of its definition shaped its behavior, what local modifications were in effect, and how state can be reconstructed or rolled back later. Materialization-heavy architectures often make these questions harder to answer because each descendant becomes an opaque bundle of assembled state. Lineage is flattened. Shared inheritance is hidden. Explanation becomes an exercise in forensic reconstruction.

Aethon preserves a cleaner record. A descendant reference can indicate exactly which canonical definition it descends from, what memory layers were in scope, what overlays individualized it, and what restrictions were applied at creation time. Because inherited structures are explicit, the system can reconstruct not only local state but also the path by which that state was acquired. This is useful for debugging, but it is even more useful for auditability.

Isolation also improves under this model when it is implemented well. Reference-based inheritance does not imply indiscriminate sharing. On the contrary, it allows scope to be expressed precisely. Descendants can inherit only those layers and capabilities that are appropriate for their role. Local writes can remain local through copy-on-write semantics. Sensitive user-specific overlays can remain distinct from broader enterprise memory. In other words, Aethon allows sharing where it is safe and useful while preserving boundaries where they matter.

Versioning is another important operational dimension. Enterprises need to evolve definitions over time without losing the ability to reproduce prior behavior. Because Aethon separates stable definitions from descendant references, it supports version-aware lineage naturally. A descendant can continue to point to the definition version from which it originated, even as newer versions are introduced. This makes rollback, comparative evaluation, and incident analysis more tractable. It also allows organizations to test new definitions in limited scopes without rewriting the identities of all existing descendants.

There is also a strategic enterprise implication. Many organizations want AI capability embedded inside their existing products, domains, and control surfaces rather than exposed only through a closed assistant experience. That means such entities must be instantiable in large numbers, close to the workflows and identities of the business. Aethon's model is well aligned with that direction because it treats creation as a cheap infrastructural operation rather than a heavyweight custom build.

Of course, operational control is not guaranteed by abstraction alone. It still requires policy enforcement, observability tooling, access controls, and disciplined operations. But architecture determines what those controls can see and how cleanly they can be applied. By retaining explicit references, layers, and lineage rather than collapsing everything into opaque clones, Aethon gives those mechanisms a more faithful picture of the system they are trying to regulate.

\section{Limitations and Open Questions}

Aethon provides a compelling framework for instantiation, but it does not eliminate all complexity. In some respects, it relocates complexity from one stage of the lifecycle to another. That relocation is often worthwhile, but it should be understood clearly.

The first major challenge is resolution discipline. Because the effective state of an instance is composed at execution time from definitions, memory layers, and local overlays, the resolver becomes a critical subsystem. If it is poorly designed, expensive, or nondeterministic, the benefits gained at instantiation time can be partially offset by execution-time uncertainty. Future implementations therefore need careful strategies for caching, partial projection, and invalidation.

A second challenge concerns memory lifecycle management. Layered inheritance and copy-on-write overlays make branching cheap, but they also introduce the possibility of a growing graph of local deltas and historical branches. Without policies for retention, compaction, archival, and garbage collection, the state graph of a long-running platform could become unwieldy. Future work should therefore explore practical strategies for pruning or consolidating overlays while preserving auditability and rollback where needed.

A third issue is developer ergonomics. Reference-based systems are semantically richer than flat object systems, but they can also be harder to inspect if the tooling is weak. Developers need to understand effective state, not just local overlays. They need to see inherited layers, precedence rules, lineage relationships, and scope boundaries. Without strong visualization and introspection tools, the conceptual elegance of Aethon could become operational opacity.

There is also the question of external side effects. Many real systems operate not only over internal memory but over external systems of record. A descendant may create CRM updates, write documents, send messages, or trigger automations. These side effects do not naturally inherit or compose in the same way internal state does. Aethon can help represent the identities that produce such effects, but the external systems may still require separate notions of transactionality, rollback, and provenance.

Another open question concerns workload diversity. The greatest benefits of Aethon are likely to appear where large numbers of descendants share substantial inherited structure and diverge only modestly. There may be workloads in which descendants differ so strongly, or external tool latency dominates so completely, that the instantiation savings are less decisive. This does not invalidate the framework, but it suggests that its comparative advantage will vary by domain.

Finally, there is the question of standardization. If reference-based instantiation becomes more common, what abstractions should be standardized across platforms? Should definitions, overlays, and lineage metadata have interoperable formats? Should there be explicit protocol surfaces for scoped inheritance and resolution? These questions matter if the broader ecosystem is to move toward portable, inspectable infrastructure rather than siloed proprietary mechanisms.

\section{Conclusion}

The infrastructure challenge of the agentic era is not only how to run powerful models. It is how to make stateful, contextual, auditable AI agents operationally real at scale. Current systems often approach this challenge using materialization logic inherited from prior software paradigms. They treat each new instance as a heavyweight object that must be assembled, hydrated, and bound before it can act. That model works tolerably well at small scale, but it becomes increasingly misaligned with the requirements of personalized, dynamic, multi-agent production systems.

Aethon proposes a different primitive. It treats an instance as a reference-based, compositional identity over stable definitions, layered memory, and local overlays. By doing so, it makes instantiation constant-time with respect to inherited structure and allows memory growth to track actual divergence rather than raw instance count. More importantly, it creates an architectural vocabulary in which specialization, branching, and lineage are natural parts of the system rather than expensive afterthoughts.

The implications of this shift are broad. It makes rich multi-agent orchestration more practical. It makes personalization less resource-intensive. It improves the substrate for auditability, reproducibility, and access control. It allows developers to design workflows according to semantic clarity instead of infrastructural fear. It suggests that the next generation of runtime platforms may need to look less like model-serving wrappers and more like operating substrates for spawnable, stateful execution identities.

There is more work to be done. Efficient resolution, lifecycle management, external side-effect reconciliation, and developer tooling all remain active areas for refinement. Yet the central insight appears durable. Much of what defines an agent instance does not need to be rebuilt every time the system creates one. If the platform can preserve stable structure, explicit lineage, and controlled local divergence, then agents~\cite{ref36} can become far cheaper and more scalable than current materialization-heavy architectures imply.

If the software industry is moving toward a world in which every product, workflow, and user may interact with specialized fleets of AI agents, then a reference-based replication primitive is not merely attractive. It may be foundational.

\end{document}